\documentclass{article}

\usepackage{microtype}
\usepackage{graphicx}
\usepackage{subfigure}
\usepackage{booktabs} % for professional tables

\usepackage{hyperref}

\usepackage{multirow}
\usepackage{amsmath}  
\usepackage{soul,color}

\usepackage{soul}
\usepackage{xcolor}

\def\figurePath{figures/}
\def\myfigure#1#2{\begin{figure}[t]\centering\includegraphics*[width = \linewidth]{\figurePath#1}
\vspace{-20pt}\caption{#2}\label{fig:#1}\vspace{-5pt}\end{figure}}
\def\mycfigure#1#2{\begin{figure*}[ht]\centering\includegraphics*[clip, width = \linewidth]{\figurePath#1}\vspace{-6pt}\caption{#2}\label{fig:#1}\vspace{-6pt}\end{figure*}}

\def\myfigurehere#1#2{\begin{figure}[h]\centering\includegraphics*[width = \linewidth]{\figurePath#1}\vspace{-8pt}\caption{#2}\label{fig:#1}\vspace{-8pt}\end{figure}}

\def\myfiguretop#1#2{\begin{figure}[t]\centering\includegraphics*[width = \linewidth]{\figurePath#1}\vspace{-8pt}\caption{#2}\label{fig:#1}\vspace{-8pt}\end{figure}}

\def\mysection#1#2{\section{#1}\label{sec:#2}}

\def\mysubsection#1#2{\subsection{#1}\label{sec:#2}}

\newcommand{\refTbl}[1]{Table~\ref{tbl:#1}}

\definecolor{unsurecolor}{rgb}{1,.85,.7}
\definecolor{changedcolor}{rgb}{.85,1,.7}

\soulregister\ref7
\soulregister\cite7
\soulregister\citeetal7
\soulregister\etal0
\soulregister\eg0
\soulregister\scene1
\soulregister\refTbl1

\DeclareGraphicsExtensions{.pdf,.ai,.psd,.jpg}
\newif\ifdraft
\draftfalse

\ifdraft
   \newcommand{\kostas}[1]{{\color{blue} {[\bf Kosta: #1]}}}
   \newcommand{\vitto}[1]{{\color{red} {[Vitto: #1]}}}
   \newcommand{\ricardo}[1]{{\color{magenta} {[Ricardo: #1]}}}
\else
  \newcommand{\kostas}[1]{}
   \newcommand{\vitto}[1]{}
   \newcommand{\ricardo}[1]{}
\fi

\newcommand{\shapenetwork}{G}
\newcommand{\appearancenetwork}{F}

\usepackage[accepted]{icml2021}

\icmltitlerunning{ShaRF: Shape-conditioned Radiance Fields from a Single View}

\begin{document}

\twocolumn[
\icmltitle{ShaRF: Shape-conditioned Radiance Fields from a Single View}

% It is OKAY to include author information, even for blind
% submissions: the style file will automatically remove it for you
% unless you've provided the [accepted] option to the icml2021
% package.

% List of affiliations: The first argument should be a (short)
% identifier you will use later to specify author affiliations
% Academic affiliations should list Department, University, City, Region, Country
% Industry affiliations should list Company, City, Region, Country

% You can specify symbols, otherwise they are numbered in order.
% Ideally, you should not use this facility. Affiliations will be numbered
% in order of appearance and this is the preferred way.
\icmlsetsymbol{equal}{*}

\begin{icmlauthorlist}
\icmlauthor{Konstantinos Rematas}{google}
\icmlauthor{Ricardo Martin-Brualla}{google}
\icmlauthor{Vittorio Ferrari}{google}
\end{icmlauthorlist}

\icmlaffiliation{google}{Google Research}

\icmlcorrespondingauthor{Konstantinos Rematas}{krematas@google.com}

% You may provide any keywords that you
% find helpful for describing your paper; these are used to populate
% the "keywords" metadata in the PDF but will not be shown in the document
\icmlkeywords{Neural Rendering}

\vskip 0.3in
]

% this must go after the closing bracket ] following \twocolumn[ ...

% This command actually creates the footnote in the first column
% listing the affiliations and the copyright notice.
% The command takes one argument, which is text to display at the start of the footnote.
% The \icmlEqualContribution command is standard text for equal contribution.
% Remove it (just {}) if you do not need this facility.

% \printAffiliations{}
\printAffiliationsAndNotice{}  % leave blank if no need to mention equal contribution
% \printAffiliationsAndNotice{\icmlEqualContribution} % otherwise use the standard text.

\begin{abstract}
We present  a method for estimating neural scenes representations of objects given only a single image. The core of our method is the estimation of a geometric scaffold for the object and its use as a guide for the reconstruction of the underlying radiance field. Our formulation is based on a generative process that first maps a latent code to a voxelized shape, and then renders it to an image, with the object appearance being controlled by a second latent code. During inference, we optimize both the latent codes and the networks to fit a test image of a new object.
The explicit disentanglement of shape and appearance
allows our model to be fine-tuned given a single image. We can then render new views in a geometrically consistent manner and they represent faithfully the input object. Additionally, our method is able to generalize to images outside of the training domain (more realistic renderings and even real photographs).
Finally, the inferred geometric scaffold is itself an accurate estimate of the object's 3D shape.
We demonstrate in several experiments the effectiveness of our approach in both synthetic and real images.
\end{abstract}

\mysection{Introduction}{intro}

% General 3D undestanding, single image, implicit functions
Understanding and reconstructing the intrinsic properties of a 3D scene, such as 3D shape and materials of objects, is a longstanding fundamental problem in computer vision~\cite{Marr1982}. The typical process for these tasks starts with a collection of multiple views of a scene and continues with the application of an algorithm to extract information such as the geometry and appearance \cite{Hartley2004, yu1999inverse}.
There are several representations for geometry such as voxels or meshes; similarly for appearance, there are colored voxels~\cite{SeitzD99}, BRDFs~\cite{Matusik2003}, etc.
Recently, there has been a paradigm shift due to the success of deep learning methods.
Firstly, the 3D scene properties can be estimated from a single image~\cite{choy20163d, drcTulsiani17, Fan_2017_CVPR, eigen2014depth,shu2017neural}.
This removes the requirement of multiple views and enables the application of a method to a wider range of scenarios.
Secondly, in some recent works the 3D scene properties are modeled with implicit representations~\cite{park2019deepsdf, onn19, mildenhall2020nerf, Oechsle2019ICCV}. These representations overcome specific shortcomings that traditional representations face (e.g. space discretization for voxels, requirement for accurate geometry for textured meshes).

% How we do it
In this paper we propose a generative method tailored for inverse rendering from a single image, using both explicit and implicit representations. In our method, the geometry and appearance of an object in the rendered image are controlled by two networks. The first network maps a latent code to an explicit voxelized shape. The second network estimates implicitly the radiance field around the object (color and volumetric density for any point) using the estimated shape as a {\em geometric scaffold}, together with a second latent code that controls appearance. Hence, the radiance field is {\em conditioned} on these two factors.
The final image is rendered by casting rays towards the scene and accumulating the color and densities to pixel values.
After training the networks, our model can input a new test image and estimate its geometric and appearance properties by re-rendering it: we optimize for the latent codes and fine-tune the network parameters so the rendered image matches the input.
At this point, our model is ready to render novel views of the test object, as both the implicit and explicit representations within our model are aware of its 3D shape.

% Why we do it
By using an explicit geometric representation for an object, we guide the appearance reconstruction to focus on its surface. The empty volume around an object does not provide useful information. Moreover, during inference on a test image we only have one view. Therefore we cannot rely on multi-view consistency for accurate shape/appearance reconstruction, as done in previous work~\cite{mildenhall2020nerf}, and our geometric scaffold compensates for that.

Our disentangled shape and appearance representation improves performance for novel view synthesis in several scenarios.
When trained and evaluated on the ShapeNet-SRN dataset~\cite{sitzmann2019srns}, our method outperforms previous works on both metrics PSNR and SSIM (except the concurrent work pixelNeRF~\cite{yu2020pixelnerf}). Here we use the standard setting of rather low-fidelity renderings with simple lighting.
Moreover,  we show that the same model, trained on these simple renderings, generalizes to other appearance domains:
(1) more realistic renderings of ShapeNet objects with complex lighting and higher resolution; and
(2) photos of real objects in the Pix3D dataset~\cite{pix3d}.
On this dataset we demonstrate better rendering quality than pixelNeRF.
Finally, our method also produces 3D objects reconstructions from a single image, achieving strong performance on par with a recent method~\cite{onn19} on two ShapeNet classes.

In summary, our contributions are:
(1) a new model to represent object classes that enables reconstructing objects from a single image,
(2) a new representation that combines an intermediate volumetric shape representation to condition a high fidelity radiance field,
and (3) optimization and fine-tuning strategies during inference that allow estimating radiance fields from real images.

\mysection{Related Work}{related}

\vspace{-2mm}
\paragraph{Scene Representation.}
To represent scenes, many works rely on geometric primitives inspired by Computer Graphics that respect the 3D nature of the world, like points, meshes, and voxel volumes. ~\citep{qi2017pointnet} reason about point clouds using permutation-invariant set networks.
Meshes can be generated by deforming a static mesh with fixed topology using a graph neural network~\cite{wang2018pixel2mesh}, modeling variable-length sequences~\cite{nash2020polygen}, or using space-partitioning functions~\cite{chen2020bspnet}.
Discretized voxel grids are a popular representation that can be naturally processed by CNNs~\cite{Brock2016GenerativeAD}, although are limited in resolution due to large memory requirements.
Surfaces can also be modeled implicitly using signed distance functions~\cite{park2019deepsdf} or occupancy volumes~\cite{chen_cvpr19, onn19}.
Other approaches have used less geometrically explicit representations, like learned latent spaces for scenes~\cite{Eslami1204} and faces~\cite{Karras2019cvpr}. Such learned latent spaces can contain subspaces with geometric properties, like viewpoint changes~\cite{harkonen2020ganspace}.
In our work, we predict voxel grids as explicit geometric scaffolds that guide the estimation of scene appearance.

\vspace{-5mm}
\paragraph{Neural Rendering.}
Scene representations are often accompanied by a renderer, which is often composed of one or multiple neural networks, a process called \emph{neural rendering}~\cite{tewari2020state}. 
Neural rendering is a nascent field with an array of diverse techniques, each specific to a particular scene representation.
Several works handle the differentiation through occlusion and visibility for meshes~\cite{li2018differentiable} and point clouds~\cite{wiles2020synsin}.
Other approaches use deferred neural rendering~\cite{thies2019deferred} and learn neural latent textures on proxy geometry to synthesize realistic views.
Other approaches raymarch through the scene accumulating latent codes to finally synthesize color values~\cite{sitzmann2019srns}. Most recently, Neural Radiance Fields (NeRF)~\cite{mildenhall2020nerf} use volumetric rendering to synthesize highly realistic novel views of a scene. NeRF has been extended to generative modeling~\cite{Schwarz2020NEURIPS} by adding conditioning on a latent code and adding adversarial losses.
Our work combines a voxelized representation that conditions a neural radiance field thereby enabling better disentanglement between shape and appearance.
This is related to \cite{Riegler2020FVS, riegler2020stable}, which use estimated geometry from multiview stereo to guide the image synthesis. However, they require several calibrated images for any new scene, in contrast to our method that uses only a single view during inference.

\vspace{-5mm}
\paragraph{3D Reconstruction.}
Classical 3D reconstruction is typically performed using stereo matching between two~\cite{scharstein2002taxonomy} or multiple calibrated views~\cite{Seitz2006}. 
A recent wave of learning-based methods have demonstrated the ability to reconstruct an object from a single RGB image, albeit in rather simple imaging conditions~\cite{onn19, Fan_2017_CVPR, wang2018pixel2mesh, choy20163d, NiemeyerDVR, DIBR19}.
Particular related are works which perform reconstruction through optimization of the latent or explicit representation by backpropagating through the rendering function~\cite{NiemeyerDVR, DIBR19}.
Meta-learning approaches can be used to accelerate the optimization process~\cite{sitzmann2019metasdf}. When modeling a latent space of scenes or objects, direct regression into that latent space can be performed~\cite{zhu2016generative}. Furthermore, cues from multiple observations can be averaged in that latent space~\cite{sitzmann2019srns}. \citep{insafutdinov2018unsupervised} further learns about object shape without direct camera pose supervision.
Our method reconstructs objects from a single image by optimizing shape and appearance latent codes and fine-tuning the neural renderer.

\vspace{-5mm}
\paragraph{Concurrent Work.}
The field of neural rendering is moving very fast, as shown by the number of arXiv preprints appearing in late 2020~\cite{dellaert2020neural}. GRF~\cite{trevithick2020grf} and pixelNeRF~\cite{yu2020pixelnerf} use image-based rendering and extract 2D features on a single or few input images to compute a neural radiance field.
Instead, our method uses a latent shape representation that provides a geometric scaffold that enables shape and appearance disentanglement.
Moreover, pixelNeRF operates in view space, which makes it difficult for the user to specify a desired new view (see Sec.~\ref{ss:pixelNeRF})
GANcraft~\cite{gancraft_arxiv} is also using explicit geometry (voxels) to render realistic images of block worlds, but it requires the scene voxels annotated with their semantic labels as an input.
IBRNet~\cite{wang2021ibrnet} combines image-based rendering with neural radiance fields and can render new scenes without re-training, but it requires multiple views of the same scene as an input.
The work of~\cite{devries2021unconstrained} estimates radiance fields given multiple views of indoor scenes using latent features in a 2D grid representing a floorplan. However, they have not demonstrated  the ability to estimate a radiance field from a single image.

\mysection{Background}{background}
\label{sec:background}

Our goal is to synthesize realistic images of an object from any viewpoint and at any resolution given a {\em single view} as input (an image with its camera parameters).
We tackle this problem by estimating the underlying neural radiance field~\cite{mildenhall2020nerf} and then render it from new views. A radiance field is a function that maps a 3D point and a 3D viewing direction to an RGB color and a density value.
We choose this representation because
(1) it provides an implicit 3D structure to the observed scene that allows consistent rendering from different viewpoints, and
(2) it can accurately represent scenes at arbitrary resolutions.

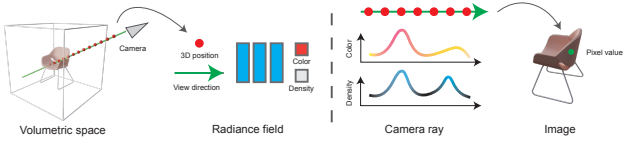
\myfigure{ray_rendering_v2}{Rendering of radiance fields. For any 3D point, a radiance field maps its 3D position, together with a view direction, to an RGB and density value. Then for a camera ray, color and density values are accumulated to deliver a final pixel color.}

\vspace{-3mm}
\paragraph{Radiance fields.}
More formally, a radiance field is a continuous function $F$ that inputs the position of a 3D point $\mathbf{p} \in R^3$ and
a viewing direction $\mathbf{d} \in R^3$,
and outputs a color value $\mathbf{c}$ and the volume density $\sigma$ at the point $\mathbf{p}$:
\begin{equation}
    F(\mathbf{p}, \mathbf{d}) \Rightarrow \mathbf{c}, \sigma
    \label{eq:vanilla_radiance_field}
\end{equation}

The radiance field can be rendered to an image pixel seen by a camera with parameters $K$ by casting a ray $r(t): \mathbf{o} + t\mathbf{d}$ from the camera center $\mathbf{o}$ towards the scene and passing through the given pixel on the image plane with ray direction $\mathbf{d}$. Using the volume rendering equations~\cite{max1995optical}, the estimated color $\hat{C}(r)$ for ray $r$ is:
\begin{equation}
    \hat{C}(r) = \int_{t_n}^{t_f} T(t)\cdot \sigma(t) \cdot \mathbf{c} (t) dt
    \label{eq:nerf_integral}
\end{equation}
where
\begin{equation}
    T(t) = \exp \left( -\int_{t_n}^{t_f} \sigma(s) ds \right )
\end{equation}
The values $\mathbf{c}(t)$ and $\sigma(t)$ are the color and volume density at point $t$ on ray $r$, and they are estimated by the radiance field $F$. The bounds $t_n$ and $t_f$ represent the nearest and farthest point of the integration along the ray and depend on the scene/camera arrangement.

In NeRF~\cite{mildenhall2020nerf},
 the above integrals are estimated with numerical quadrature: a set of random quadrature points $\{t_k\}_{k=1}^{K}$ is selected between $t_n$ and $t_f$ with stratified sampling and the final color can be estimated as:
\begin{equation}
    \hat{C}(r) = \sum_{k=1}^{K} \hat{T}(t_k) \cdot \xi(\sigma(t_k) \cdot \delta_k) \cdot \textbf{c} (t_k)
    \label{eq:volume_render}
\end{equation}
\begin{equation}
    \hat{T}(t) = \exp \left( -\sum_{k=1}^{k-1} \sigma(t_k) \cdot \delta_k \right )
    \label{eq:nerf_opacity}
\end{equation}
where $\xi(x) = 1- \exp(-x)$ and $\delta_k = t_{k+1}-t_k$ is the distance between two points on the ray.

In NeRF, the radiance field $F$ is estimated using a neural network.
The network is trained using a set of images $I$ of the same object/scene by optimizing the following loss:
\begin{equation}
    \sum_{i} \sum_{j \in I_i}||C(r_{ij}) - \hat{C}(r_{ij}) || ^2
    \label{eq:nerf_loss}
\end{equation}
where $C(r_{ij})$ is the ground truth color of ray $j$ passing through a pixel in image $i$. Each image is calibrated, i.e. the camera parameters $K$ are given.

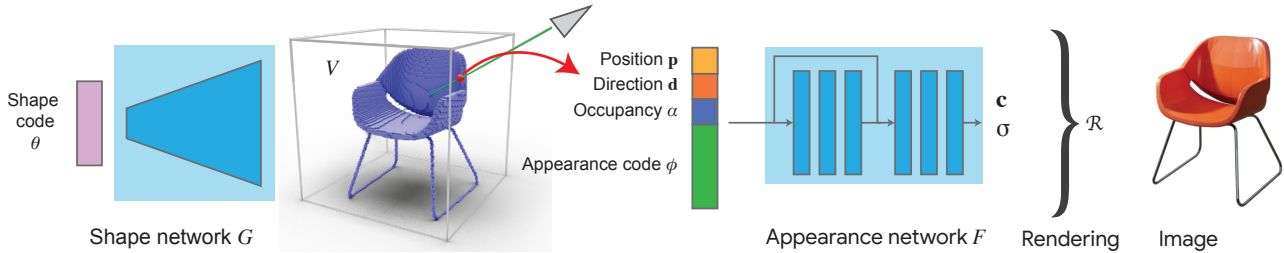
\mycfigure{overview}{Overview of our generative process: a shape code is mapped to a 3D shape and an appearance code controls its appearance.
}

The function $F$ uses the multiview consistency among the calibrated images $I$ to implicitly capture the 3D nature of the underlying scene. However, one major limitation is that each function $F$ implicitly embeds the geometry and appearance of a particular scene, and thus a separate neural network model needs to be trained for every new scene. Again, a large number of calibrated images is required to accurately train $F$ and hence to render new views of a scene.

\vspace{-3mm}
\paragraph{Rendering a new view.}
The radiance field $F$ can be used to render a new view specified by camera parameters $K'$ as follows \cite{mildenhall2020nerf}. 
For each pixel, we cast a ray from the camera center towards the scene, passing through the pixel in the image plane.
Then, for each ray we apply the process in Eq.~\eqref{eq:nerf_integral}-\eqref{eq:nerf_opacity} to determine the color of its corresponding pixel (Fig.~\ref{fig:ray_rendering_v2}).
The points $\mathbf{p}$ and direction $\mathbf{d}$ in those equations are all derived from the camera $K'$ with basic geometry.
We refer to this process of generating a new image $I'$ given the radiance field $F$ and camera parameters $K'$ as $\mathcal{R}(F, K') \Rightarrow I'$.

\section{Our method}

In this section we present our generative framework for image synthesis and how it is used for single image radiance field estimation during inference.

\mysubsection{Generative Neural Rendering}{generative_neural_rendering}

The goal of our method is to estimate the radiance field of an object from a single image so that we can render novel views.
We approximate the image formation process by a generative neural rendering process that is conditioned on two latent variables, one that controls the shape of the object and another that controls its appearance (Fig.~\ref{fig:overview}).
We now describe a full pass through the model.

First, a {\em shape network} $\shapenetwork$  maps a latent code $\theta$ into a 3D shape represented as a voxel grid $V \in R^{128^3}$. Each voxel $i$ contains a scalar $\alpha_i \in [0, 1]$ indicating its occupancy: $\shapenetwork(\mathbf{\theta}) \Rightarrow V$.
The voxel grid specifies a continuous volumetric area in world coordinates where the object exists. 

Second, we estimate a radiance field, i.e. the color $\mathbf{c}$ and density $\sigma$ of any 3D point $\mathbf{p}$ inside this area, using an {\em appearance network} $\appearancenetwork$ which extends~\eqref{eq:vanilla_radiance_field}:
\begin{equation}
    \appearancenetwork(\mathbf{p}, \mathbf{d}, \alpha_p, \mathbf{\phi}) \Rightarrow \mathbf{c}, \sigma
    \label{eq:app_net}
\end{equation}
This extended radiance field is now conditioned on two elements:
(1) the voxelized shape $V$ produced by the shape network, via the occupancy value $\alpha_p$ at point $\mathbf{p}$; and
(2) a latent code $\phi$ that controls the appearance of the object.

Finally, we can synthesize a new view of the object with the rendering process $\mathcal{R}$ of the radiance field (described at the end of Sec.~\ref{sec:background}). Note that all operations between the two networks $\shapenetwork$ and $\appearancenetwork$ are fully differentiable.

Our generative formulation brings several advantages.
It sets up the intrinsic properties of the object (shape and appearance) in such a way that estimating them is effective and precise. By using an explicit geometric scaffold, we condition the network $\appearancenetwork$ to estimate radiance fields for a specific 3D shape.
In particular, we guide the network to estimate proper color values on the surface of the object by indicating the occupancy $\alpha_p$ and the appearance latent code $\phi$ for a particular object). 
Additionally, during inference, we cannot rely on multi-view constraints as we are given a single test image. In this situation, the geometric scaffold provides valuable 3D information to steer the radiance field estimation towards the object surface.
Finally, conditioning the appearance network on the shape scaffold produced by $\shapenetwork$ enforces a disentanglement between the shape and appearance latent spaces by construction, which is beneficial when generalizing to different domains (Sec.~\ref{sec:exp_shapenet_realistic}).

\mysubsection{Shape Network}{shape_network}
We use a discretized voxel grid for representing the shape scaffold.
Voxel representations integrate naturally with convolutional architectures and the chosen resolution provides a good balance between geometric details and memory requirements (our $128^3$ resolution is sufficient for capturing even fine details of a single object).
In addition, for any point $\mathbf{p}$ inside the extent of the voxel grid, we can estimate its occupancy value $\alpha_p$ efficiently using trilinear interpolation. This process is differentiable, allowing the  communication between the shape network $\shapenetwork$ and the appearance network $\appearancenetwork$. Moreover, the scaffold provides a strong geometric signal to the appearance network, both by considering it as an input via $\alpha_p$, and by enabling the sampling of more points on the object surface for Eq.~\eqref{eq:volume_render}.

\vspace{-3mm}
\paragraph{Architecture.}
The network consists of a series of fully connected layers, followed by a series of 3D convolutional blocks (with ReLU and batch normalization layers; details in the supplementary material).
This network maps a shape latent code $\theta$ to a voxelized shape $V$.%\ricardo{Update if we remove Fig 4}

\vspace{-3mm}
\paragraph{Training.}
As training data, we use the 3D objects of a ShapeNet class (e.g. chairs or cars) and each instance has its own shape latent code $\theta$.
During training we optimize both the network $\shapenetwork$ and these latent codes $\theta$ (see Fig.~\ref{fig:overview}), akin to the Generative Latent Optimization (GLO) technique~\cite{bojanowski18a}.
The loss consists of three parts.
First, the weighted binary cross entropy loss between a predicted $\hat{V}$ and a ground truth voxel grid $V$~\cite{Brock2016GenerativeAD}.
Second, we use a symmetry loss on the voxels, as we assume the objects are left/right symmetric.
Finally, we incorporate a voxel-to-image projection loss by projecting the estimated voxels to two random views $j$ and comparing it with the corresponding object silhouette $S_j$.
The overall loss is:
\begin{align*} 
&\frac{1}{|V|} \sum_{i\in V} \gamma \alpha_i \log{\hat{\alpha}_i}+ (1-\gamma) (1-\alpha_i) \log{(1-\hat{\alpha}_i)}\\
&+ w_{sym}||\hat{V} - \hat{V}_{sym}||^2 + w_{proj}\sum_{j \in\{1,2\}}||\mathcal{P}_j(\hat{V})-S_j||^2
\end{align*}
where $\hat{\alpha}_i$ is the occupancy at voxel $i$,
$w_{sym}$ and $w_{proj}$ are the weights for the symmetry and projection loss respectively,
$\gamma$ is a weight to penalize false negatives,
and $\mathcal{P}_j$ is the differentiable projection of the object silhouette on the random view $j$.
The latter operation is similar to Eq.~\eqref{eq:volume_render}, without the $\mathbf{c}$ factor.

Note that while we use 3D voxelized shapes for training, this is in practice not a big limitation as they can be easily computed for synthetic datasets, or estimated using traditional 3D reconstruction techniques for real datasets with multiple views.
Note how standard neural rendering methods typically train from large number of multiple views~\cite{sitzmann2019srns, mildenhall2020nerf, DupontICML20, Tatarchenko16, Schwarz2020NEURIPS}.

% \myfigurehere{decomposition_v2}{We estimate a geometric scaffold for the object, which helps to generate an accurate volumetric appearance and density.}

\mysubsection{Appearance Network.}{appearance_network}

Our appearance network $\appearancenetwork$ models a radiance field~\eqref{eq:app_net} similar to the one of  NeRF~\eqref{eq:vanilla_radiance_field}, but extended to include additional conditioning inputs:
(a) the occupancy value $\alpha_p$ at $\mathbf{p}$, as estimated by the shape network $\shapenetwork$,
and 
(b) the appearance latent code $\phi$ controlling the appearance of the object.

\vspace{-3mm}
\paragraph{Architecture.}
The appearance network $\appearancenetwork$ has similar architecture to NeRF. It consists of a series of fully connected layers (followed by ReLU) that map the above input to an RGB color $\mathbf{c}$ and a density value $\sigma$.

\vspace{-3mm}
\paragraph{Training.}
As training data, we use the same 3D objects as in Sec.~\ref{sec:shape_network}.
For each object we render $N=50$ views, and for every view we sample random rays passing through its pixels. Note that an appearance latent code $\phi$ represents a single object, therefore the code is shared among all its views. The final data consists of all the rays from every image, together with the corresponding shape latent code $\theta$, the appearance latent code $\phi$, and the ground truth pixel color for each ray.

As in Sec.~\ref{sec:shape_network}, we use GLO to optimize the network $\appearancenetwork$ together with the appearance latent codes $\phi$ for each training object.
For every ray $r$ in the training set, we sample points with stratified sampling ~\cite{mildenhall2020nerf}.
For every sampled 3D point $\mathbf{p}$, we estimate its occupancy value $\alpha_p$ from the voxel grid $\shapenetwork(\theta)(\mathbf{p}) = V(\mathbf{p})$.
These are input to the appearance network $\appearancenetwork$, together with the viewing direction $\mathbf{d}$ and the appearance code $\phi$ for this object.
The network outputs a color $\mathbf{c}$ and a density $\sigma$ for $\mathbf{p}$, 
which are then accumulated along the ray $r$ as in Eq.~\eqref{eq:volume_render} giving the final color for the pixel.
We train the appearance network $\appearancenetwork$ by minimizing the loss \eqref{eq:nerf_loss}, comparing this final color with the ground truth.

The training of $\appearancenetwork$ depends on the occupancy values of the voxel grid $V$. While we can use the $V$ output by the shape network $G(\theta)$ directly, we achieve higher rendering quality by first pre-training $\appearancenetwork$ using the ground truth voxels as $V$, and then fine-tuning with the estimated $V$ from $\shapenetwork(\theta)$.

% \myfigurehere{voxel_network_v2}{Generative model for 3D shape estimation.}

\subsection{Inference on a test image}
\label{subsec:optimization}

Our model is essentially a renderer with two latent codes $\theta,\phi$ that control the shape and appearance of an object in the output image. The model is differentiable along any path from the latent codes to the rendered image, therefore it can be used for the inverse task: given a test image $I$ and its camera parameters $K$, we can reconstruct the latent codes and then use them for synthetizing a new view of that object.

\vspace{-3mm}
\paragraph{Estimating the latent codes.}
We can estimate the latent codes for a test image $I$ by:
\begin{equation}
    \arg\min_{\theta, \phi} || \mathcal{R}(\appearancenetwork,K) - I||^2 +||\shapenetwork(\theta) - \shapenetwork(\theta)_{sym}||^2
    \label{eq:optimization}
\end{equation}
where $\mathcal{R}$ is the rendering process from Sec.~\ref{sec:background} ('Rendering a new view').
Note that $\appearancenetwork$ from Eq.~\eqref{eq:app_net} depends on $\phi$, and on $\theta$ via its arguments $\alpha_p$ (Sec. \ref{sec:generative_neural_rendering}).
The first term of this loss measures the difference between $I$ and its re-rendered version. Hence, minimizing this term means finding the latent codes $\theta,\phi$ that lead to rendering the test image.
The second term encourages the voxel grid estimated by $\shapenetwork$ to be symmetric.

\vspace{-3mm}
\paragraph{Jointly optimizing latent codes and networks.}
In practice, the above objective is difficult to optimize.
The implicit assumptions are that the input image $I$ can be represented by latent codes sampled from their respective latent space, and that it is possible to discover them using gradient descent methods.
When the test image is substantially different than the training set, then the appearance latent code $\phi$ cannot properly express the colors in it.
Moreover, the voxel grid output by the shape network $\shapenetwork$ tend to lack thin details.
These two phenomena lead to blurry renderings by the appearance network $\appearancenetwork$. 

However, we observed that minimizing the objective~(\ref{eq:optimization}) also over parameters of the shape and appearance networks $\shapenetwork,\appearancenetwork$, in addition to the latent codes $\theta,\phi$, results in more detailed shape reconstructions and more accurate renderings, even when the test image is from a different domain than the training set. This corresponds to fine-tuning the networks $\shapenetwork,\appearancenetwork$
beyond the parameters found during training (Sec. \ref{sec:shape_network} and \ref{sec:appearance_network}),
allowing to overfit to this particular test image.

In practice, we devised through experimentation that the best results can be achieved by a two-stage optimization procedure, which we describe below.

Importantly, notice how the objective~\eqref{eq:optimization} does not require any annotations for the test image apart from the camera parameters (e.g. no ground truth 3D shape).
Hence, optimizing it w.r.t. the network parameters is a valid operation even at test time.

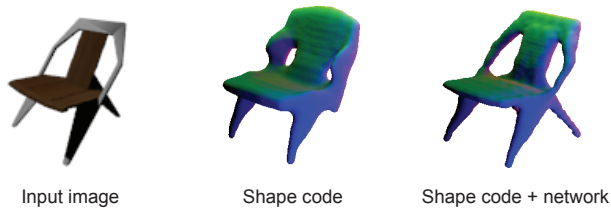
\myfigurehere{shape_estimation_v2}{Shape reconstruction by (a) optimizing only the shape code, (b) together with fine-tuning the shape network. }

\paragraph{Stage 1: Shape code $\theta$ and network $\shapenetwork$.}
In the first stage, we focus on optimizing the shape code and network parameters in order to get an accurate shape estimation for the object in the test image. To achieve this, we use the appearance network $\appearancenetwork$ as a neural renderer that we can backpropagate through to measure the image reconstruction loss~\eqref{eq:optimization}.
Concretely, we keep $\appearancenetwork$ fixed and optimize the shape code $\theta$, shape network $\shapenetwork$ and appearance code $\phi$, to achieve the best image $I$ reconstruction~\eqref{eq:optimization}.
In other words, we ask the shape code $\theta$ and network $\shapenetwork$ to produce an accurate shape so that when it is rendered with appearance code $\phi$ and network $\appearancenetwork$, it reproduces the input image $I$.

The fine-tuning of the network $\shapenetwork$ is a necessary element for accurate reconstruction, as the optimization of only the shape code $\theta$ results in approximate reconstructions and missing geometric details (Fig.~\ref{fig:shape_estimation_v2}).
In the experiment section we further evaluate the performance of different optimization strategies.

\vspace{-3mm}
\paragraph{Stage 2: Appearance code $\phi$ and network $\appearancenetwork$.}
In Stage 2, we optimize for the appearance code $\phi$ and network $\appearancenetwork$, while keeping the shape code and network fixed to the output of Stage 1. Again, our goal is for the rendering $\mathcal{R}$ to be as close as possible to input image $I$ by minimizing~\eqref{eq:optimization}.
In other words, we ask the appearance code $\phi$ and network $\appearancenetwork$ to render the shape $V=\shapenetwork(\theta)$ (estimated in Stage 1) so as to match the input image $I$.

The fine-tuning of appearance network  $\appearancenetwork$ is also important, as it results in more accurate  reconstructions of the input image (Fig.~\ref{fig:app_recon_v2}). This is particularly noticeable when the test image $I$ is significantly different than the training data, in which case the optimization of the appearance code $\phi$ is insufficient. We present a thorough evaluation of this setting in Sec.~\ref{sec:exp_shapenet_realistic}.

\vspace{-3mm}
\paragraph{Rendering a new view $K'$.}
So far, we have found the shape and appearance codes and updated networks that enable to re-render the test image $I$.
In order to render a new view $I'$ specified by camera parameters $K'$, we use the updated latent codes and networks in $\mathcal{R}(\appearancenetwork,K')$.
Note that our optimizations and fine-tuning avoids trivial solutions, e.g. when the test image $I$ is re-rendered accurately but new views are wrong.
Our method achieves this by
(1) incorporating prior knowledge of the object class using a learned latent shape and appearance models ($\shapenetwork$, $\appearancenetwork$);
(2) adding a symmetry prior in the form of a symmetry loss (Eq. ~\eqref{eq:optimization});
(3)
customizing $\appearancenetwork$ to focus on the surface of the object by conditioning it on the geometric scaffold.

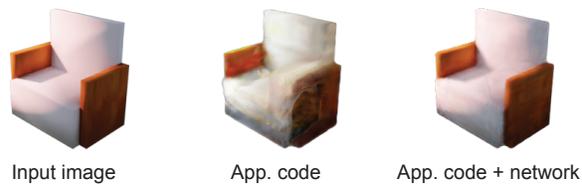
\myfiguretop{app_recon_v2}{Image reconstruction by (a) optimizing only the appearance code, (b) together with fine-tuning the appearance network.}

\mysection{Experiments}{experiments}

We extensively evaluate our approach for novel view synthesis given a single image on three datasets:
(1) an existing benchmark based on ShapeNet (ShapeNet-SRN~\cite{sitzmann2019srns}, Sec.~\ref{sec:exp_shapenet_srn});
(2) an updated ShapeNet test set with images rendered more realistically  and at higher resolution (dubbed ShapeNet-Realistic, Sec.~\ref{sec:exp_shapenet_realistic});
(3) the Pix3D~\cite{pix3d} dataset, with photographs of real objects (Sec.~\ref{sec:exp_pix3D}).
As additional experiments,
we
(4) demonstrate our method can operate without being given camera parameters and compare to pixelNeRF (Sec.~\ref{ss:pixelNeRF});
(5) investigate the effects of exploiting symmetry at test time (Sec.~\ref{ss:symmetry});
(6) report results for 3D shape reconstruction on two ShapeNet classes (Sec.~\ref{sec:exp_3Drec}).
Additional results can be found in the supplementary material.

\vspace{-3mm}
\paragraph{Metrics.}
We evaluate with the standard image quality metrics PSNR and SSIM~\cite{ssim}.

\vspace{-3mm}
\paragraph{Technical details.}
In all experiments we use positional encoding~\cite{mildenhall2020nerf} of a 3D point $\mathbf{p}$ with 6 frequencies.
The appearance network $\appearancenetwork$ has two residual blocks of fully connected layers with width 256. We use 128 rays per image and 128 stratified samples per ray, plus 128 importance samples estimated by the geometric scaffold. The latent codes are initialized from a normal distribution and all optimizations are performed with Adam~\cite{KingmaB14}. During inference, we optimize the latent codes and networks for 1000 iterations.

\subsection{Variants of our method}
\label{sec:exp_variants}
Our method is highly modular and adjustable.
We investigate four variants of our method, which differ in how the object shape is estimated and how it interacts with the appearance network.

\vspace{-3mm}
\paragraph{Variant 1: Conditional NeRF.}
This is a simpler version of the appearance network $\appearancenetwork$ in~\eqref{eq:app_net}, which only inputs the position of a point ${\mathbf p}$ and the appearance code $\phi$. Hence, 
we train the network with GLO, optimizing both the network parameters and the appearance codes.
This variant does not depend on any explicit geometry (no shape scaffold); elements of both the object appearance and geometry are entangled in the only latent code $\phi$.
This network can be seen as a variant of GRAF~\cite{Schwarz2020NEURIPS} with only one (optimized) latent code as an input.

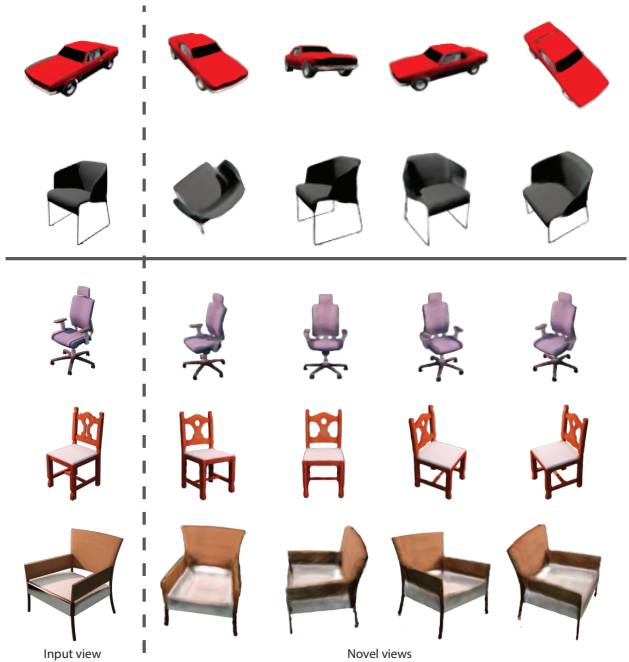
\myfiguretop{shapenet_realistic}{Results on ShapeNet-SRN (top two rows) and ShapeNet-Realistic (other rows).}

\vspace{-3mm}
\paragraph{Variant 2: ShapeFromNR.}
The second variant corresponds to our description in Sec.~\ref{subsec:optimization}:
the appearance network $\appearancenetwork$ has the same form as in~\eqref{eq:app_net}, and both latent codes as network parameters are fit to the test image (NR stands for neural rendering).

Here the geometric scaffold is estimated by optimizing~\eqref{eq:optimization}, i.e. the reconstruction of the test image. Recall that in doing so, the appearance network helps by acting as a renderer in Stage 1.
Then, in Stage 2 the shape network guides the appearance network in the rendering process and thus support fitting $\phi,\appearancenetwork$ to the test image.
Hence, the two networks interact fully.

\begin{table}
\begin{center}
\begin{tabular}{ lcc } 
 Variant & code-only &  code+network\\ 
\hline
V1 Conditional NeRF & 22.12 / 0.90 & 22.05 / 0.91\\
V2 ShapeFromNR  & 23.37 / 0.92 & 23.31 / 0.92\\
V3 ShapeFromMask  & 22.94 / 0.91 & 22.98 / 0.91\\
\hline
V4 ShapeFromGT  & 25.59 / 0.94 & 25.65 / 0.94\\
\hline
\end{tabular}

\end{center}
\vspace{-5pt}
\caption{Analysis of four variants of our method on Chairs from ShapeNet-SRN. Results in form PSNR/SSIM.
}
\label{tab:shapenet_srn}
\end{table}

\begin{table}
\begin{center}
\begin{tabular}{ cccc } 
 &  &  PSNR & SSIM\\ 
\hline
\multirow{5}{*}{\rotatebox[origin=c]{90}{Chairs}}
    % &GRF~\cite{grf2020}& 21.25& 0.86\\
    &GRF~\cite{trevithick2020grf}& 21.25& 0.86\\
    &TCO~\cite{Tatarchenko16}& 21.27& 0.88\\
    &dGQN~\cite{Eslami1204}& 21.59& 0.87\\
    &SRN~\cite{sitzmann2019srns}& 22.89& 0.89\\
    &ENR~\cite{DupontICML20}& 22.83& -\\
    &pixelNeRF~\cite{yu2020pixelnerf}& \textbf{23.72}& 0.91\\
    &Ours (ShapeFromNR)& 23.37& \textbf{0.92}\\
\hline
\multirow{3}{*}{\rotatebox[origin=c]{90}{Cars}}
    & SRN~\cite{sitzmann2019srns} & 22.25 & 0.89 \\
    &ENR~\cite{DupontICML20}& 22.26& -\\
    &pixelNeRF~\cite{yu2020pixelnerf}& \textbf{23.17}& \textbf{0.90}\\
    &Ours (ShapeFromNR)& 22.53& \textbf{0.90}\\
\hline
\end{tabular}
\end{center}
\vspace{-5pt}
\caption{Comparison with other methods on Cars and Chairs of ShapeNet-SRN. pixelNeRF is a concurrent work.}
\label{tab:shapenet_srn_comparison}
\end{table}

\vspace{-3mm}
\paragraph{Variant 3: ShapeFromMask.}
In this variant the geometric scaffold is instead estimated from a given segmentation mask of the object during Stage 1. Concretely, we optimize the shape network $\shapenetwork$ and the latent code $\theta$ to fit the mask  by minimizing the projection $\mathcal{P}$ of the estimated shape with the given mask (see 'Training' in Sec.~\ref{sec:shape_network}).
In this variant the shape and appearance networks interact less: the shape network is estimated first based on the mask alone, and then it is fed to the appearance network for Stage 2.
As the ShapeNet-SRN dataset does not provide masks, we estimate them automatically using background subtraction, as the backgrounds are uniformly white.

\vspace{-3mm}
\paragraph{Variant 4: ShapeFromGT.}
Here we directly provide the ground truth object shape as $V$ to the appearance network (determining the input occupancy $\alpha$). There is no shape network in this variant.
This can be seen as an upper bound on the performance on our method, where the appearance network can rely on a perfect indication of where the object surface is, and thus it can more easily render it properly based on the visible parts in the test image.

\subsection{Novel View Synthesis on ShapeNet-SRN}
\label{sec:exp_shapenet_srn}

\paragraph{Settings.}
We use the same experimental setup and data as \cite{sitzmann2019srns}.
The dataset consist of 6591 Chairs and 3514 Cars that are split for training, validation and testing.
For training, each object is rendered with simple lighting from 50 different viewpoints to images of $128^2$ resolution, with the cameras lying on a sphere around the object.
For testing, the objects are rendered from 251 views on an archimedean spiral with the same illumination and resolution as training.

At test time, we first fit the model to one of the 251 views as in Sec.~\ref{subsec:optimization} (plays the role of the test image $I$), with a procedure that depends on the variant (Sec.~\ref{sec:exp_variants}).
Then, we render each of other 250 views as described at the end of Sec.~\ref{subsec:optimization} (they play the role of `new views' requested by the user).
Finally, we compare the rendering to the ground truth image to evaluate performance.

\vspace{-3mm}
\paragraph{Results.}
Tab.~\ref{tab:shapenet_srn} show the performance of the four method variants on the Chair class.
First, all variants 2,3,4, which include some form of shape scaffold, outperform variant 1. This supports the main contribution of our paper.
Interestingly, ShapeFromNR performs somewhat better than ShapeFromMask.
This is likely because the training and test images come from the same distribution (same illumination, resolution, etc). Hence the appearance network can accurately guide the shape reconstruction.
As expected, ShapeFromGT achieves the best performance, as the ground truth shape is the best guide for the appearance network.

In the columns of Tab.~\ref{tab:shapenet_srn}, we compare across another axis: optimizing the appearance latent code only, or together with the appearance network.
We see that, in this setting, the difference is minor.
This is also due to the training and test images having similar characteristics, so the test images can be fit well by optimizing only the appearance latent codes.
We show qualitative results in Fig.~\ref{fig:shapenet_realistic}.

\vspace{-3mm}
\paragraph{Comparison to other works.}
We compare our ShapeFromNR method to recent results on this dataset in Tab.~\ref{tab:shapenet_srn_comparison} (numbers taken from the original papers).
We outperform all previous methods on both classes and for both metrics, with an especially visible improvement on Chairs.
Our results are also on par with the concurrent method pixelNeRF~\cite{yu2020pixelnerf} on Chairs, while being somewhat below on Cars.

\subsection{Novel View Synthesis on ShapeNet-Realistic}
\label{sec:exp_shapenet_realistic}

\paragraph{Settings.}
We now perform novel view synthesis on a different synthetic test set.
We render each Chair from the test split of ShapeNet-SRN using a path tracer with complex illumination (environment map and local lights, ambient occlusion) to 20 views of resolution $256^2$ pixels. These images are more realistic than the original test set of ShapeNet-SRN and they have very different appearance statistics (shadows are in different locations, the color palette is more vibrant, etc.).
All model variants are still trained on the original ShapeNet-SRN training set, allowing us to investigate their ability to generalize to a different appearance domain.

\begin{table}
\begin{center}
\begin{tabular}{ lcc } 
 Variant & code-only &  code+network\\ 
\hline
V1 Conditional NeRF & 21.86 / 0.86 & 22.91 / 0.89\\
V2 ShapeFromNR  & 21.68 / 0.88 & 22.07 / 0.89\\
V3 ShapeFromMask  & 22.68 / 0.88 & \textbf{23.26 / 0.90}\\
\hline
V4 ShapeFromGT  & 24.84 / 0.91 & 25.65 / 0.92\\
\hline
\end{tabular}
\end{center}
\vspace{-5pt}
\caption{Evaluation on test Chairs from ShapeNet-Realistic. Here the test images are rendered more realistically than the training set (ShapeNet-SRN). Results in form PSNR/SSIM.}
\label{tab:shapenet_realistic}
\end{table}

\vspace{-3mm}
\paragraph{Results.}
Tab.~\ref{tab:shapenet_realistic} presents results.
First, two of the variants with a shape scaffold (V3,V4) outperform V1 (Conditional NeRF), which confirms it helps also in this more challenging setting.
Moreover, Conditional NeRF's renderings are blurry and show floating artifacts, especially in the presence of thin structures (chair legs in Fig.~\ref{fig:comparisons}).
To quantify this further, we computed the perceptual metric LPIPS~\cite{zhang2018perceptual} on this dataset (lower is better):
0.109 for ShapeFromMask vs 0.115 for Conditional NeRF.

In this setting, ShapeFromMask delivers better results than ShapeFromNR. As the test images are substantially different from the training set, the segmentation mask provides a better signal for shape reconstruction than neural rendering from a appearance network trained in a different domain. 

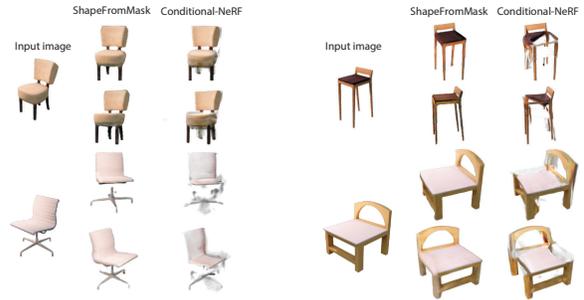
\myfigure{comparisons}{Comparison between our ShapeFromMask and Conditional NeRF on ShapeNet-Realistic.}

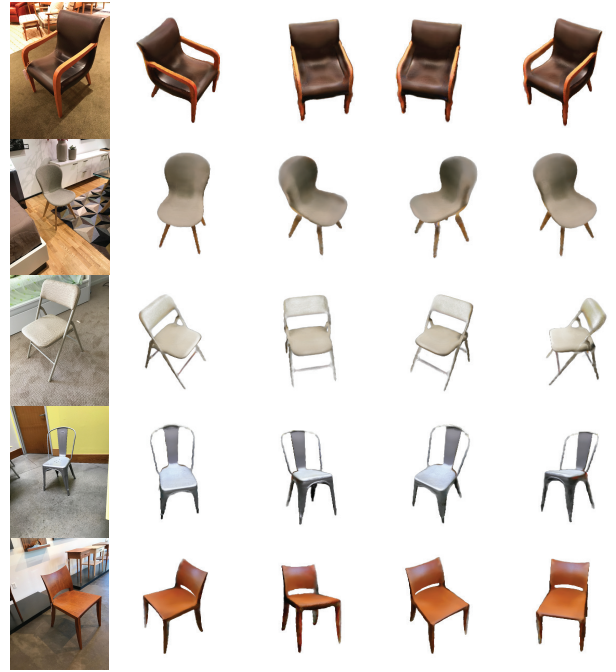
\myfigure{pix3d}{Results on Pix3D using our ShapeFromMask variant.\vspace{-3mm}
}
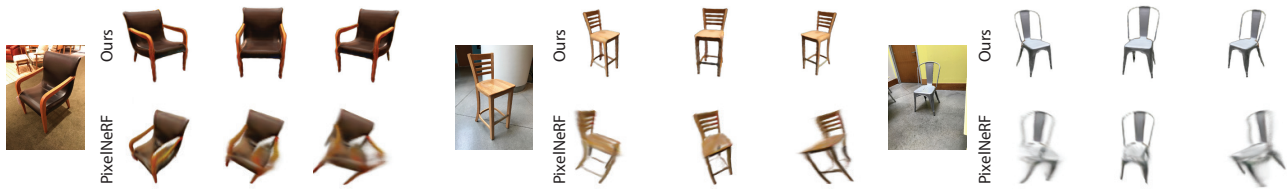
\mycfigure{est_camera}{Comparison with pixelNeRF on Pix3D (here our method automatically estimates camera parameters).}

Finally and most importantly, by comparing across the columns of Tab.~\ref{tab:shapenet_realistic}, we see that optimizing the code+network substantially improves the accuracy of all variants, compared to code-only.
In this setting the difference is much larger than in Sec.~\ref{sec:exp_shapenet_srn}, as
optimizing also the networks enables to bridge the gap between the training and test domains.
We show qualitative results in Fig.~\ref{fig:shapenet_realistic} using ShapeFromMask with code+network optimization.

\vspace{-3mm}
\subsection{Novel View Synthesis on Pix3D}
\label{sec:exp_pix3D}
We also apply our ShapeFromMask method to real photographs from Pix3D~\cite{pix3d},
using the provided masks as input (Fig.~\ref{fig:pix3d}).
The model is still trained on the simple ShapeNet-SRN renderings, again demonstrating generalization to a new domain.

\vspace{-3mm}
\subsection{Comparison to pixelNeRF}
\label{ss:pixelNeRF}

Nearly all novel view synthesis methods based on neural rendering, e.g. SRN~\cite{sitzmann2019srns}, GRF~\cite{trevithick2020grf}, DVR~\cite{NiemeyerDVR}, NeRF~\cite{mildenhall2020nerf} assume known camera parameters.
The very recent PixelNeRF~\cite{yu2020pixelnerf} circumvents this requirement by operating in view space.
However, this makes is difficult for a user to specify a particular new viewpoint for synthesis (e.g. a frontal view of a chair),
because it needs to be expressed relative to the internal view space coordinate system (which varies for each input image depending on the depicted object pose).

However, we can easily incorporate automatic estimation of the camera parameters at test time (3D rotation, 3D translation) into our method.
We first retrieve the closest rendered image from the training set (ShapeNet-SRN) using $L_2$ distance on HOG features.
Then, we use the camera parameters of the retrieved image as an initialization and we optimize them for the test image together with the latent codes $\phi,\theta$ as in Eq.~\eqref{eq:optimization}.
This simple approach results in just $13.8^{\circ}$ rotation error and $0.14$ blender units translation error in ShapeNet-SRN.
Note that the parameters correspond to a canonical space so a user can specify a novel view directly.

We now compare to PixelNeRF using the public code, trained on the same dataset as ours (ShapeNet-SRN) and applied to Pix3D chairs. We make sure to place the chairs a white background based on their mask, as the pixelNeRF code expects them in this format.
We compare to our ShapeFromMask variant, also without given camera pose, in Fig.~\ref{fig:est_camera}.
The images rendered by pixelNeRF are blurrier than our method's, and thin structures exhibit ghosting. 
There are also clear artifacts from perspective distortion as pixelNeRF operates in view space: the assumed camera frustum does not match the actual one, making the object appear stretched when seen from a new view.
Finally, operating in view space makes the objects seem slanted: the new view can only be specified as a relative transformation, reducing the user ability to control the viewpoint.

\vspace{-3mm}
\subsection{Using symmetry.}
\label{ss:symmetry}
Many man-made objects exhibit symmetry along one axis and we can take advantage of that, similarly to other works~\cite{wu2020unsup3d, Yaocvpr20}.
Having the symmetric object in a canonical camera frame allows us to use the mirrored input image during shape/appearance optimization at test time as well.
We used this for our qualitative results in Fig.~\ref{fig:shapenet_realistic} and \ref{fig:pix3d}, as it yield slightly better renderings. 
Also quantitatively, the performance with symmetry is only slightly higher than without ($0.91/23.51$ vs $0.90/23.26$ SSIM/PSNR on ShapeNet-Realistic ShapeFromMask).
We stress that all quantitative results in Tab.~\ref{tab:shapenet_srn},~\ref{tab:shapenet_srn_comparison},~\ref{tab:shapenet_realistic} are without exploiting symmetry at test time.

\vspace{-3mm}
\subsection{Shape reconstruction}
\label{sec:exp_3Drec}
We additionally evaluate the 3D reconstruction performance of the shape network $\shapenetwork$ for the ShapeFromMask variant in the standard ShapeNet setting of~\cite{choy20163d} using the evaluation protocol of Occupancy Networks~\cite{onn19}. 
Our method performs on par with Occupancy Networks for Chair (ours 0.49 vs 0.50 IoU) and better for Car (ours 0.77 vs 0.74 IoU), though this comparison is approximative as our method is class-specific.
While we provide this limited evaluation as an indication that our reconstructions are good, the focus of our work is to use the 3D reconstructions as a means towards accurate renderings.

\vspace{-3mm}
\mysection{Conclusion}{conclusion}
We present a method for estimating a radiance field from a single image using both explicit and implicit representations.
Our generative process builds a geometric scaffold for an object and then then uses for estimating the radiance field. By inverting the process, we recover the explicit and implicit parameters and use them for synthesizing novel views. We show state-of-the-art results on a standard novel view synthesis dataset and we demonstrate generalization to images that differ significantly from the training data.

\bibliography{sharf}
\bibliographystyle{icml2021}

%%%%%%%%%%%%%%%%%%%%%%%%%%%%%%%%%%%%%%%%%%%%%%%%%%%%%%%%%%%%%%%%%%%%%%%%%%%%%%%
%%%%%%%%%%%%%%%%%%%%%%%%%%%%%%%%%%%%%%%%%%%%%%%%%%%%%%%%%%%%%%%%%%%%%%%%%%%%%%%

\end{document}